\documentclass{article}

\usepackage[preprint]{neurips_2020}

\usepackage[utf8]{inputenc} 
\usepackage[T1]{fontenc}    
\usepackage{hyperref}       
\usepackage{url}            
\usepackage{booktabs}       
\usepackage{amsfonts}       
\usepackage{nicefrac}       

\usepackage{microtype}      

\usepackage[english]{babel}
\usepackage{graphicx}
\usepackage{subcaption}
\usepackage{enumitem}
\usepackage{float}

\setcitestyle{numbers,open={[},close={]}}

\title{Formatting the Landscape: Spatial conditional GAN for varying population in satellite imagery}

\author{
  Tomas Langer \thanks{contact author: langer.tomas@yahoo.com}  \\
  Intuition Machines Inc \\
 \And
  Natalia Fedorova \\
  Max Planck Institute for Evolutionary Anthropology \\
 \AND
  Ron Hagensieker \\
  osir.io \\
  
}

\begin{document}

\maketitle

\begin{abstract}

Climate change is expected to reshuffle the settlement landscape: forcing people in affected areas to migrate, to change their lifeways, and continuing to affect demographic change throughout the world. Changes to the geographic distribution of population will have dramatic impacts on land use and land cover and thus constitute one of the major challenges of planning for climate change scenarios. In this paper, we explore a generative model framework for generating satellite imagery conditional on gridded population distributions. We make additions to the existing ALAE \citep{Pidhorskyi2020alae} architecture, creating a spatially conditional version: SCALAE. This method allows us to explicitly disentangle population from the model's latent space and thus input custom population forecasts into the generated imagery. We postulate that such imagery could then be directly used for land cover and land use change estimation using existing frameworks, as well as for realistic visualisation of expected local change. We evaluate the model by comparing pixel and semantic reconstructions, as well as calculate the standard FID metric. The results suggest the model captures population distributions accurately and delivers a controllable method to generate realistic satellite imagery.  

\end{abstract}

\section{Introduction}

Human beings are not actionless pawns in the face of climate change, they adapt to both direct and indirect pressures brought about by an increasingly unstable climate. People can either choose to leave problematic areas, migrating both locally and internationally to adapt to their circumstances, or they can stay and change their lifeways. In either case, and on top of expected demographic change, human adaptation to climate change thus reshuffles the settlement landscape \citep{Devitt2015}. As local populations ebb and flow, land use and land cover change in response. Due to this high mobility of populations, particularly given recent work on climate induced migration \citep{Rigaud2018}, one of the major challenges for planning for climate change scenarios is thinking about where people will be, and how this will change the landscape. State-of-the-art work on gridded population forecasts shows us the value of a greater geographic resolution and border oblivious approach \citep{Rigaud2018,Jones2020}. However, such forecasts still require analytic processing to evaluate their consequences for local landscapes.

In this paper, we explore the potential for generating satellite imagery conditional on population change as an in-between step for analysis, and as a means of directly visualizing forecasts in a realistic way. To do so, we employ the latest generative models from the field of unsupervised machine learning, namely generative adversarial networks \citep{Goodfellow2014}. GANs are a state-of-the-art technique for high resolution image generation, as they can generate images from random noise, also called the latent space. We refer the reader to a review article \citep{wang2020ganreview} for additional details. Generative models have been successfully applied to various high resolution datasets, mainly faces (e.g. CelebAHQ \citep{karras2018progressive}, FFHQ \citep{karras2019stylegan}), objects (e.g. Imagenet \citep{imagenet_cvpr09}), and scenes (e.g. LSUN \citep{yu15lsun}, Cityscapes \citep{cordts2016cityscapes}). However, generative models in the high resolution earth observation domain are relatively under-explored \citep{Rolnick2019}.

On top of image generation, one might wish to edit a real image via a trained generative model. For this, a mapping from image space to latent space is required. This is a natural feature of autoencoders \citep{Kingma2014ae} or flow based generative models \citep{kingma2018glow}. However, one can learn the mapping for a trained GAN model as well, which is called projection \citep{abdal2019image2stylegan,karras2020stylegan2}. In order to facilitate fast and accurate mapping, we decided to train a hybrid model based on Adversarial Latent AutoEncoder (ALAE) \citep{Pidhorskyi2020alae}, which combines a GAN and an AE by training a generator and an encoder end-to-end.

Furthermore, to explicitly control the generated population, we add a conditional input to the generator (more information in appendix \ref{gan_info_appendix}), where the input label is a pixel-level population map.

Our main contributions are:
\begin{itemize}[noitemsep,topsep=0pt]
 \setlength\itemsep{0em}
 \item Adding spatial conditioning module to the ALAE architecture
 \item Training the model on satellite + population data
 \item Visualizing population in generated images
 \item Evaluating the generative model's performance in terms of quality and the population change effect
\end{itemize}

\section{Methods}

\subsection{Data}

Two data sets are utilized over a study site of continental Central America including Mexico. We focus on Central America here as it has been identified as a region already experiencing high levels of internal and international migration due to climate change, and has been the focus of prior modelling efforts \citep{Jones2020}. However, this approach could of course be applied to any geographic region. 

Image data is derived from surface reflectance data from ESA's Sentinel-2 mission \citep{louis2016sentinel}. Sentinel-2 is a constellation of two satellites, which collect images at a 5 day revisit. The second data set is the Global Human Settlement population data (GHS-POP) for the years $2000$ and $2015$ from the European Commission's Joint Research Centre \citep{schiavina2019ghs,Florczyk2019}. Both datasets are publicly available, and  the details of how we sample the data are available in appendix \ref{sampling_appendix}.

\subsection{Model architecture}

\begin{figure}[!ht]
  \centering
  \includegraphics[height = 4cm, width = 13cm]{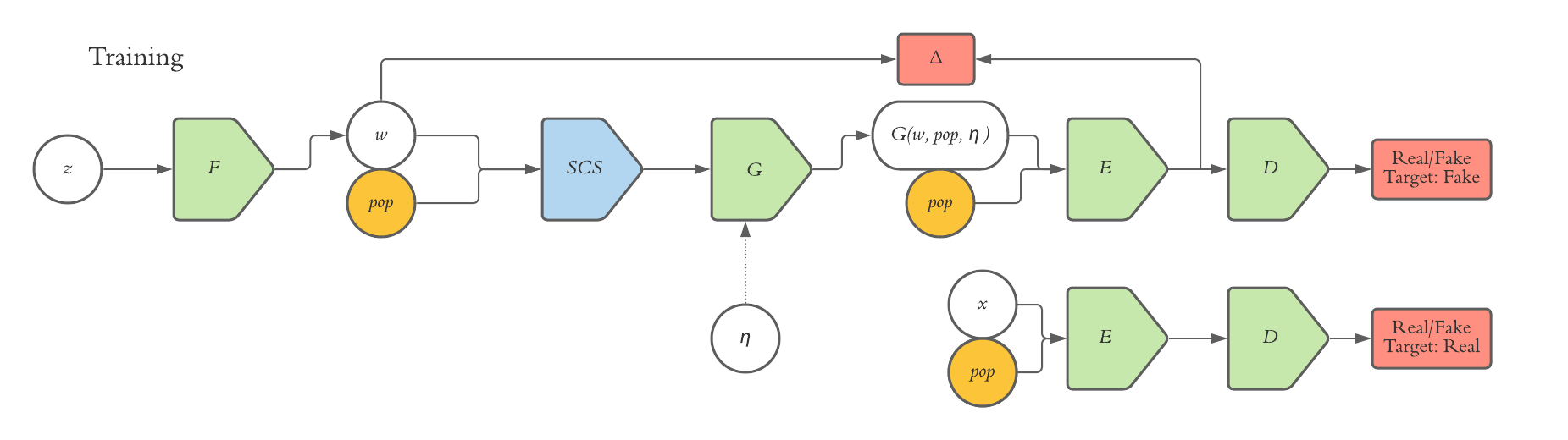}
  \caption{\textbf{SCALAE model training} where white circles represent original ALAE inputs, yellow circles are our additional inputs, green boxes are original ALAE modules, blue boxes are our modified modules, and red boxes are the losses.}
  \label{alae_new_training}
\end{figure}

We use ALAE \citep{Pidhorskyi2020alae} for the basis of our model, which is in turn based on the StyleGAN \citep{karras2019stylegan} architecture. We refer the reader to these papers for in depth description of the architecture, and focus here instead on our modifications. We adapt the ALAE codebase \citep{pidhorskyi2020alaegithub} for our training and evaluation, and our additional code and trained models can be accessed here \footnote{https://github.com/LendelTheGreat/SCALAE}.

Our training model differs from ALAE in 2 ways: the added \emph{Spatial Conditional Style} (SCS) module, and the modified encoder input. Hence, the equation 2 from the ALAE paper can be modified to equation \ref{eqn1}. All training modifications are also shown in figure \ref{alae_new_training}.

\begin{equation} \label{eqn1}
  {\textrm{G = } G \circ SCS \circ F \textrm{,  and  D = } D \circ E} 
\end{equation}

The SCS module is used to feed conditional information (population map) to the generator. The style input \emph{w} and appropriately resized population map \emph{pop} are combined together by a learned function that maps both to the same number of channels, then summed together, and finally fed into the adaptive instance normalization layers of the generator. This process can be seen in figure \ref{scs} in the appendix \ref{scs_appendix}.

The discriminator gets the conditional information (population map) as an input so that it provides a learning signal based on how the population map and the generated image fit together. This is done via a simple concatenation of the RGB and population channels. Because the ALAE discriminator is partly an encoder, the population map is concatenated with the RGB channels of either real or fake image and fed into the encoder to predict the style \emph{w}, which is then fed into the discriminator head.

\subsection{Reconstruction}

After model training, we reconstruct a real satellite image by mapping it to the latent space of the model. This is trivial thanks to the autoencoder architecture, visualized in figure \ref{alae_new_inference} in appendix \ref{scalae_recon_appendix}.

Since the population map serves as an input to the generator, we can feed in a custom population map and control the population in the reconstructed output. This process can be used to visualize how real world places would look like with an alternative population distribution, for example, a climate change induced population change.

\section{Results}

\subsection{Generation model quality}

To evaluate quality of generations we calculate the Fréchet Inception Distance (FID) \citep{heusel2018fid}, more information in appendix \ref{FID_appendix}. The overall model FID is $45.13$ (lower is better) for random generations. Our work would benefit from general benchmarking in the field, as at the moment, FID scores are only useful as a comparison across different models evaluated on the same data set. We thus report the FID score with the hopes of stimulating further comparison in the future.

In the absence of a direct comparison, we visually check the images, and as portrayed by figure \ref{random_samples} in appendix \ref{random_gen_appendix} we confirm that generations strongly resemble realistic satellite imagery with sufficient diversity.

\subsection{Reconstruction quality}

To further evaluate model quality we focus on reconstructing real satellite imagery. In this case, we give the model a real reference image as an additional input, which, as expected, improves the FID to $35.97$. Examples of reconstructions can be seen in figure \ref{reconstructions} and more in appendix \ref{reconstructions_apendix}.

\begin{figure}[!ht]
  \centering
  \includegraphics[height = 4.6cm, width = 13.8cm]{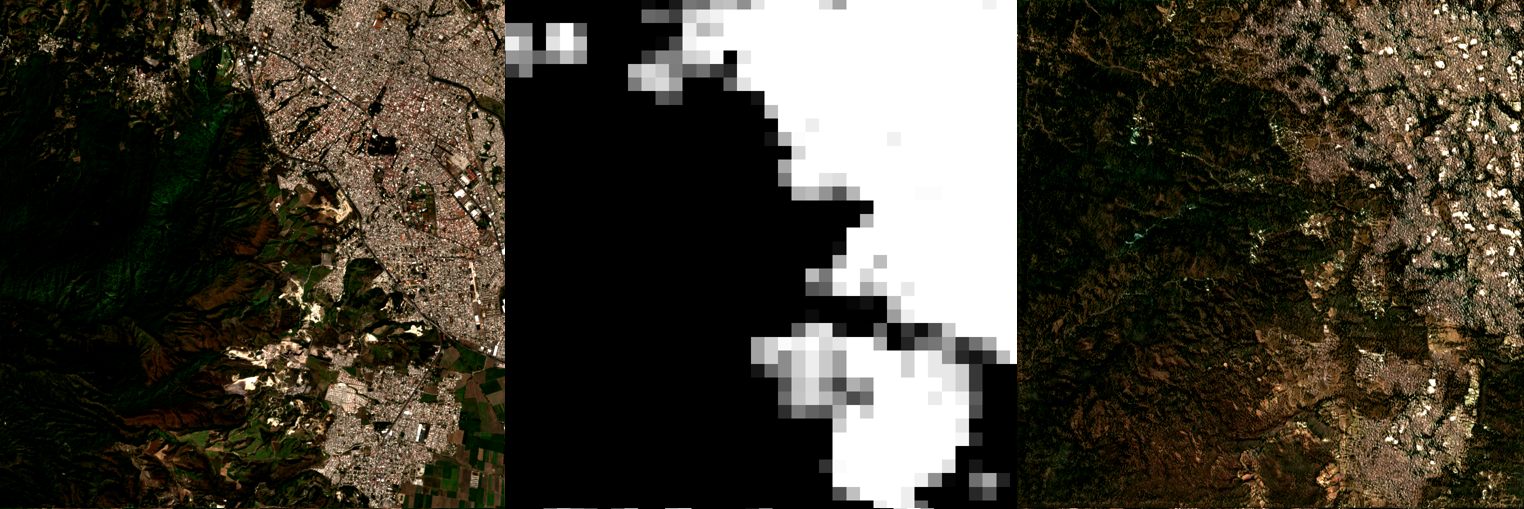}
  \caption{\emph{left}: original input image, \emph{center}: input population map, \emph{right}: generated image}
  \label{reconstructions}
\end{figure}

By generating reconstructions of real images, we create matching pairs that can be used for evaluation. We calculate the difference between the pairs of each real vs reconstructed image. The difference measure can be done on the pixel level, and on the semantic level using pretrained Inception features (same features as used in the FID score calculation). We use a standard l2 distance to compute the difference measures.

For the pixel level, the mean pixel l2 distance between pairs is $277.40$, but a frequency plot of the distribution of pixel distances for $9436$ image comparisons has a long tail, so much worse images in terms of pixel reconstruction are also possible, shown in figure \ref{pixel} in appendix \ref{hist_appendix}. The mean value of the semantic distance is $15.15$ and here a frequency plot of the distances is much more normally distributed, meaning better and worse images are equally likely (see figure \ref{semantic} in appendix \ref{hist_appendix}). The extreme tails of both of the above measure are visualized in figure \ref{best_worst} in appendix \ref{best_worst_appendix}.
    
\subsection{Population effect}

\begin{figure}[!ht]
  \centering
  \includegraphics[height = 3cm, width = 6cm]{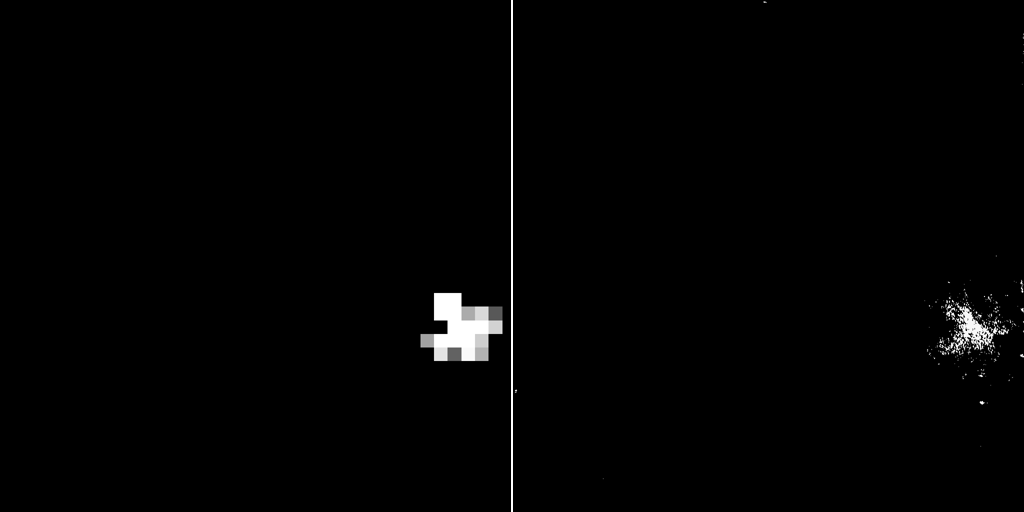}
  \caption{\emph{left}: population input, \emph{right}: pixel difference averaged over 20 styles}
  \label{pop_diff_avg}
\end{figure}

To evaluate the effect of population on reconstructed images, and thus show the efficacy of our model (i.e. how well the population conditioning works), we produce generations for varying population inputs and visualize the pixel difference between them. In figure \ref{pop_diff_avg}, we calculate the pixel difference averaged over 20 samples of style vectors \emph{w}, showing the high consistency of the model reconstruction. Thus highlighting that the population conditioning is spatially consistent. An example of this process is visualized in figure \ref{pop_diff_ex} in appendix \ref{pop_pix_dif_appendix} and in figure \ref{pop_manipulation} in appendix \ref{pop_manipulation_appendix}.

\section{Discussion and Concluding remarks}

We have created a model architecture that makes it possible to spatially condition style-based generative methods and thus to explicitly disentangle the latent space from a spatial label. We show that the population in the generated images can be manually controlled in a fine-grained manner, giving the user the ability to change population in specific parts of an image. Moreover, the encoder of our network can be used to map real images to the latent space, making it possible to not only edit fake, but also real, imagery. We believe this model could be useful for visualizing climate change related population forecasts such as those modelled in \citep{Jones2020}, as it allows practitioners and researchers to generate imagery flexibly, concretely, and with a means to characterize uncertainty.

Furthermore, the ability to map real images to the latent space opens up several image editing possibilities. We can continuously perform latent space arithmetic to create meaningful changes in the generated images, following previous GAN examples (e.g. \citep{karras2019stylegan,karras2020stylegan2,harkonen2020ganspace}). Moreover, combining latent space arithmetic with explicit population conditioning delivers more control over exactly what is generated and where. Importantly, this can be done continuously, not just to generate a static outcome, but also to interpolate between or visualize a distribution of possible outcomes.

Likewise, it is difficult to evaluate the climate change effect on population on real imagery without reference longitudinal data. This will become more possible as longitudinal satellite data collections with matching population grids become available for longer time spans. Finally, the imagery we generate can be fed directly into existing frameworks for land use and land cover analysis, without further retraining or adaptation.

\begin{ack}

First of all, we would like to thank Lucas Kruitwagen, our mentor from the NeurIPS 2020 “Tackling Climate Change with Machine Learning” workshop mentorship program, for feedback on structuring the project and positioning within wider literature. Next we are thankful to Intuition Machines Inc for providing the necessary compute resources for this project, and Tom Bishop from Intuition Machines for general feedback on the paper. Last but not least, we would like to thank Björn Lütjens, Esther Wolf, and Aruna Sankaranarayanan for fruitful discussion on the topic of generating satellite imagery and its relation to climate change.

\end{ack}

\bibliography{biblio}

\section*{Broader Impact}

We envision that our research here could be of benefit for both local and international organizations who are committed to integrating AI  
methodologies into their planning and policy workflows. However, given the complexity of the models and the infrastructure and financing required for training, there is an obvious gap in who is able to actually use these models leading to concerns of centralization.

In terms of biases, we identified several sources of bias in our method, that could lead to undesired outcomes, and should be explicitly taken into account before any direct application of this method.
First of all, there is the model bias of GANs themselves. GANs are know to suffer from mode dropping, which could results in uncommon features of the dataset being ignored by the generative model, even if they are presents in the data. This bias can be approximately quantified by measuring diversity of generations using recent methods \citep{gu2020giqa} and visualized, which means it can be evaluated in relation to particular use cases. Moreover, recent methods have made substantial improvements to mitigate this bias \citep{srivastava2017veegan,dieng2019presgan,yu2020inclusive}. Note that our method SCALAE is a hybrid that includes an autoencoder in the latent space, which partly reduces the mode dropping problem, however, proper evaluation of this phenomena is ongoing research and is left for future work.

Secondly, there are several sources of bias in the data itself. On one hand, it is the data collection bias. For example, because of cloud cover, satellite data collection usually focuses on the dry season only. On the other hand, there is some concern for leveraging path dependency biases, coming from the fact that the generated images can only reflect patterns that were observed before. It thus cannot capture new developmental trajectories. Nonetheless, there is interesting new research being done in the direction of novelty generation \citep{phdthesis}.

Finally, we do not anticipate, or recommend, that a generative approach be used in isolation for policy and planning - it is a tool to aid field professionals and ideally should be linked directly with theoretically and locally informed behavioral models, in fact, we postulate that the conditioning approach we develop here makes this imminently possible. 

\section{Appendices}

\subsection{Additional background on conditional GANs} \label{gan_info_appendix}

In addition to image generation, GANs and other methods have shown promising results on controlling the generation process. This can be done when the latent space is sufficiently disentangled, in which case, adjusting a section of the latent space results in a meaningful semantic change in the generated image. A well disentangled latent space can be learned in a completely unsupervised fashion \citep{chen2016infogan}, however, providing an explicit signal to the model, if available, results in more precise control over the generated image \citep{chen2019selfsupervised}. Therefore, our method focuses on so called conditional image generation. Conditioning, even though it requires additional labels, helps disentanglement and allows explicit content generation.

The difference between unconditional and conditional image generation is that the former generates realistic images from random noise, whereas the latter generates realistic images from a given input label. This label can come in many forms, usually image-level \citep{Mirza2014,brock2019large,Oeldorf2019loganv2} or pixel-level \citep{isola2018pix2pix,Park2019spade,Zhu2020sean} attributes, or even images from a different domain \citep{wang2018pix2pixhd,zhu2020cyclegan}. In our case, the input label is a pixel-level population map.

\subsection{Related works} \label{gan_related_appendix}

A popular section of generative models relevant to our problem are the domain transfer models \citep{isola2018pix2pix,wang2018pix2pixhd,zhu2020cyclegan}. These methods produce a function that translate images between 2 or more domains (e.g. horse to zebra, day to night, etc). In comparison with our method, they do not allow for sampling random images. Thus, our method has the advantage of being used outside of its current use case without retraining, an important point when we consider the financial and environmental costs. However, most domain transfer methods contain direct skip connections between the reference real image and the generated output, which is known to reproduce better pixel-wise reconstructions. 

Another side of work focuses on discovering meaningful latent space directions after the model was fully trained \citep{harkonen2020ganspace,shen2020interfacegan,abdal2020styleflow}. This is complimentary to our model and the exploration of these is left for future work.

\subsection{Data: Sampling information} \label{sampling_appendix}

Sampling sites for Sentinel-2 imagery were determined based on largest population increase in this time period, and a set of $9436$ tiles with extents of $1024$ by $1024$ pixels was extracted. The selected sites are shown in appendix \ref{site_aq_appendix}. Surface reflectance data of the dry season (Jan-March) was cloud-masked and averaged utilizing Google Earth Engine \citep{gorelick2017google}. Finally, GHS-POP data was reprojected into the corresponding UTM-zones of the Sentinel-2 tiles. Given the illustration purposes of this study we have not included the non-visible bands of Sentinel-2, and focus on the RGB channels only, as those are easily interpretable to the human eye. However, our method is invariant to the number of channels used and could be trivially retrained with the full multi-spectral depth of Sentinel-2.

\subsection{Sites for image acquisition} \label{site_aq_appendix}
\begin{figure}[H]
  \centering
  \includegraphics[height = 8cm, width = 12cm]{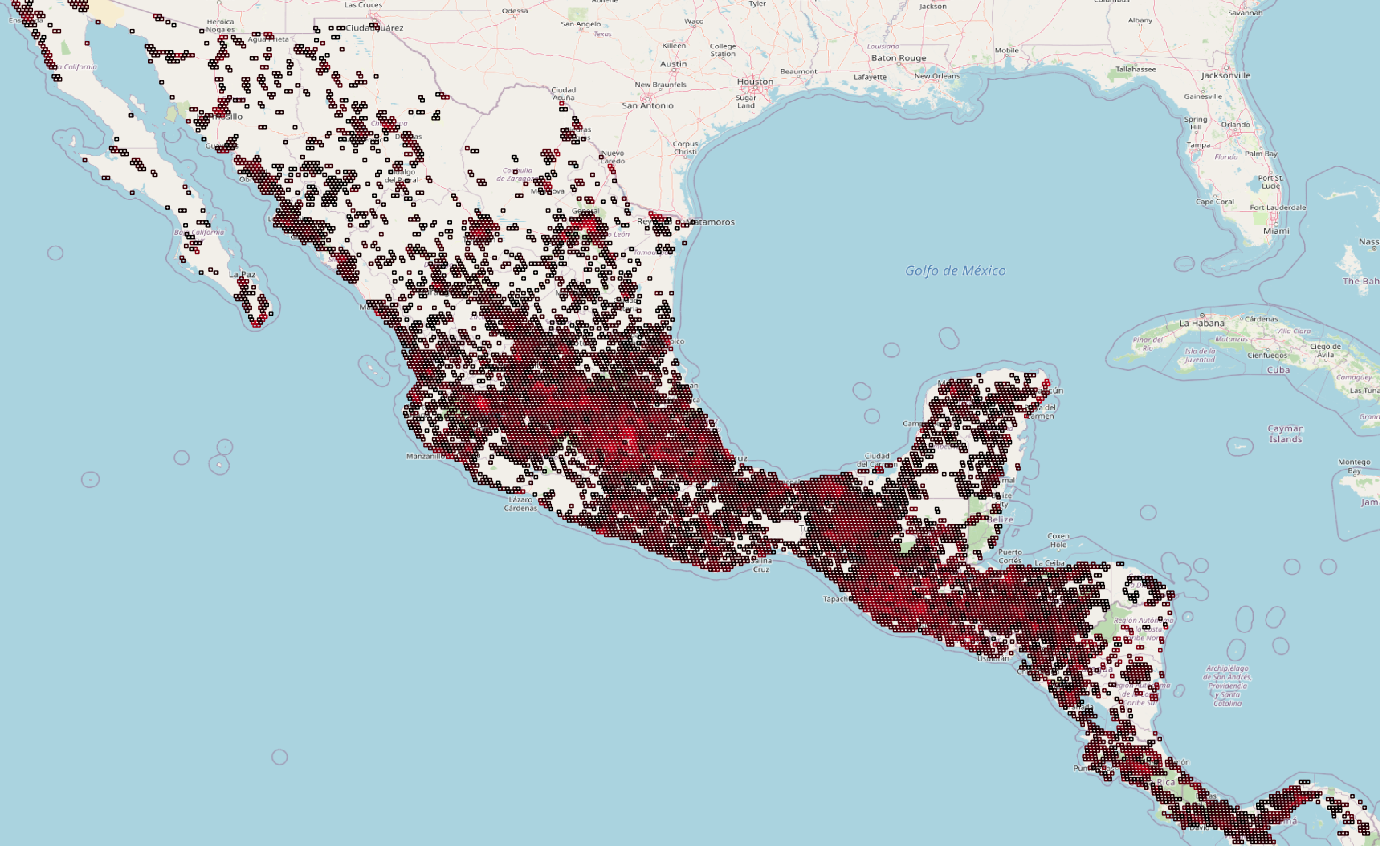}
  \caption{\textbf{Sites for image acquisition} Image acquisition based on population growth between 2000 and 2015; brighter color indicates higher population growth.}
  \label{image_acquisitions}
\end{figure}

\subsection{SCS module} \label{scs_appendix}
\begin{figure}[H]
  \centering
  \includegraphics[height = 4cm, width = 6cm]{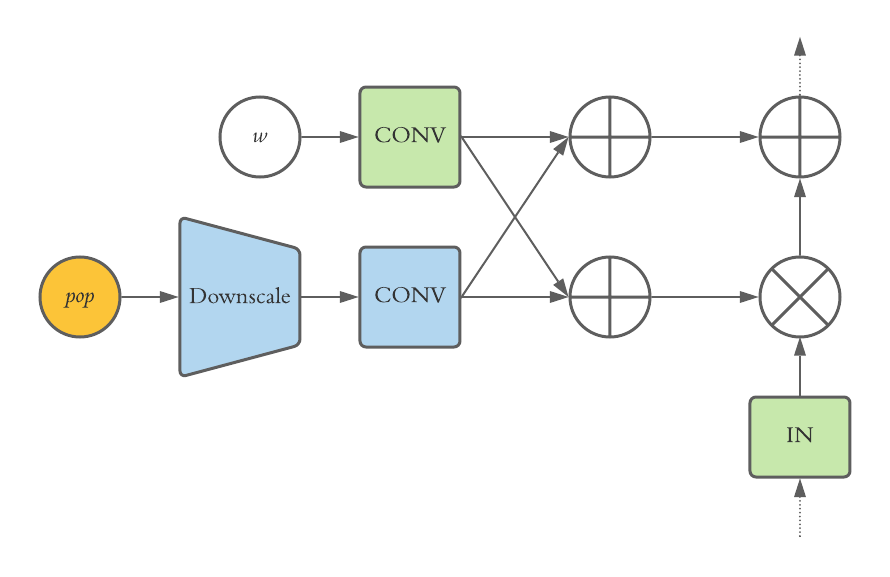}
  \caption{\textbf{SCS module} To the best of our knowledge, adding spatial conditioning to a StyleGAN-like architecture has not been explored before. \citep{Oeldorf2019loganv2} successfully add a non-spatial class conditioning into a StyleGAN by feeding a class label \emph{c} together with noise \emph{z} into the F network and subsequently into the G network. This approach however, does not take spatial dimensions into account. We take inspiration from \citep{Park2019spade,Zhu2020sean} as they use the same AdaIN \citep{huang2017adain} layers as StyleGAN, and introduce the SCS module.}
  \label{scs}
\end{figure}

\subsection{SCALAE reconstruction method} \label{scalae_recon_appendix}
\begin{figure}[H]
  \centering
  \includegraphics[height = 3.5cm, width = 10cm]{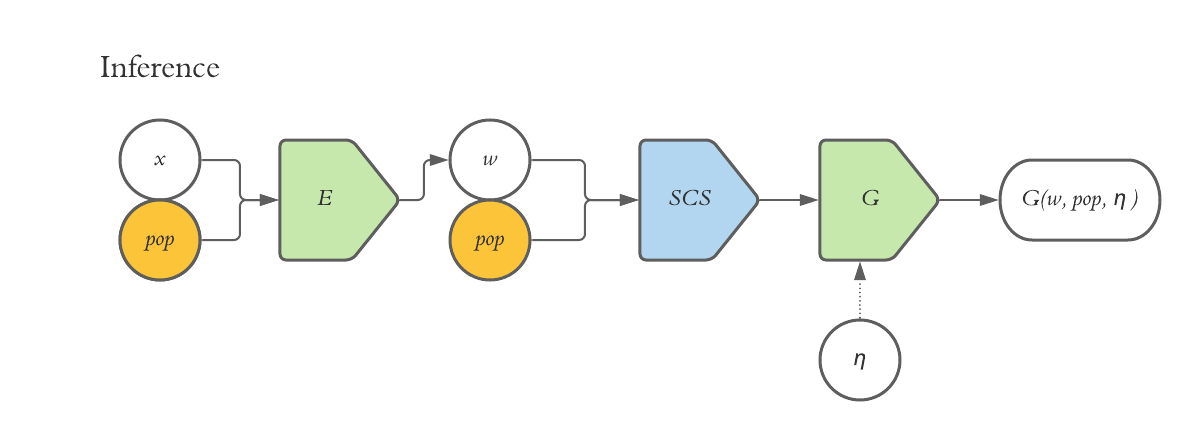}
  \caption{\textbf{SCALAE reconstruction method} where white circles represent original ALAE inputs, yellow circles are our additional inputs, green boxes are original ALAE modules, and blue boxes are our modified modules.}
  \label{alae_new_inference}
\end{figure}

\subsection{Training details} \label{training_appendix}
We train the SCALAE model end-to-end on the paired Sentinel-2 and GHS-POP data sets on 4xV100 GPUs, following the default training parameters from the ALAE codebase \citep{pidhorskyi2020alaegithub}. The Sentinel-2 imagery RGB channels are scaled between -1 and 1. The population map is log transformed and likewise scaled between -1 and 1. We train with progressive growing for 200 epochs, switching to higher resolution every 16 epochs. Our base learning rate is 0.002 and batch size 512, both adjusted accordingly with the default progressive growing schedule. The training losses remain unchanged.

Our code and all training parameters will be publicly released upon publication.

\subsection{FID details} \label{FID_appendix}

The FID score uses features from an Inception network \citep{szegedy2014inception} trained on the Imagenet \citep{imagenet_cvpr09} data set, which consists of natural photos of various objects. However, it does not include any images from satellite or other earth observation domain. We note that the domain shift in this case is not problematic because Imagenet-trained networks mostly focus on textures \citep{geirhos2019imagenettrained}, which are the main features we are trying to quantify.

\subsection{Reconstruction: distance histograms}\label{hist_appendix}
\begin{figure}[H]
  \centering
  \begin{subfigure} {.5\textwidth}
      \centering
      \includegraphics[height = 3cm, width = 5cm]{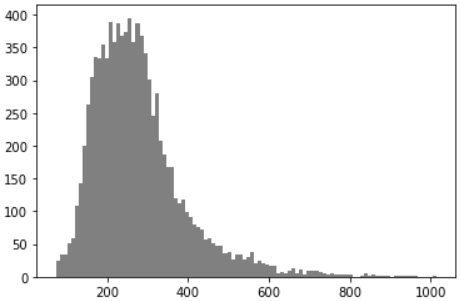}
      \caption{Pixel distance}
      \label{pixel}
  \end{subfigure}%
  \begin{subfigure} {.5\textwidth}
      \centering
      \includegraphics[height = 3cm, width = 5cm]{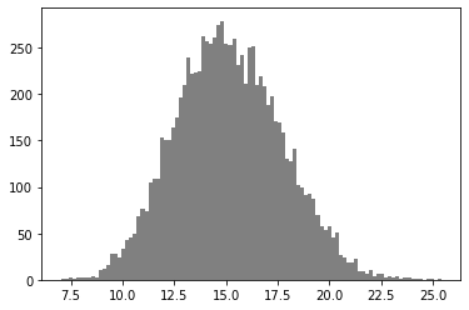}
      \caption{Semantic distance}
      \label{semantic}
  \end{subfigure}
  \caption{\textbf{Histograms of distances}}
\end{figure}

\subsection{Additional visualisations}

\subsubsection{Random generations}\label{random_gen_appendix}
\begin{figure}[H]
  \centering
  \includegraphics[height = 3.45cm, width = 13.8cm]{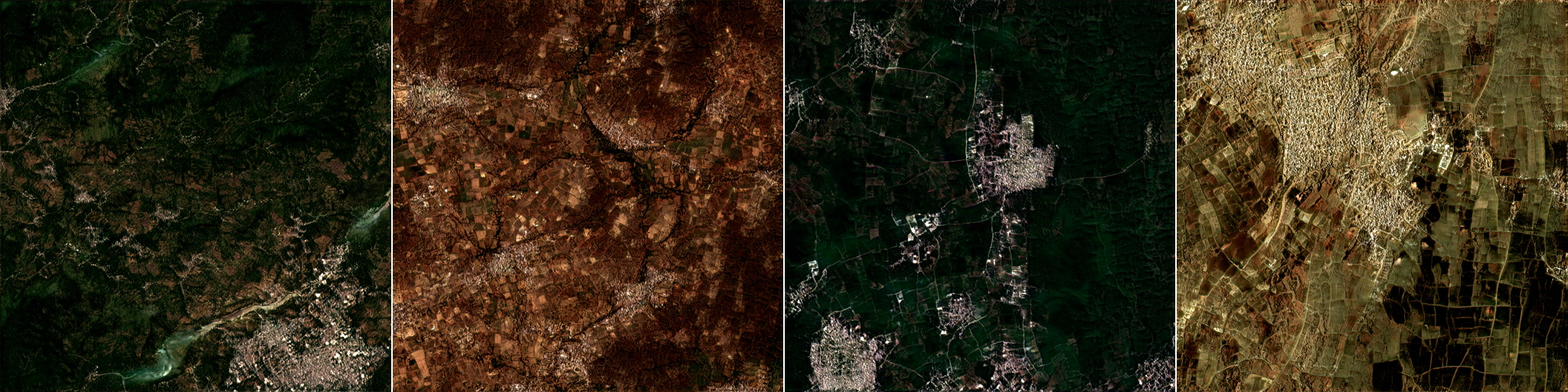}
  \caption{\textbf{Random generated samples}}
  \label{random_samples}
\end{figure}

\subsubsection{Reconstructions}\label{recon_appendix}
\begin{figure}[H]
  \centering
  \includegraphics[height = 15cm, width = 11.25cm]{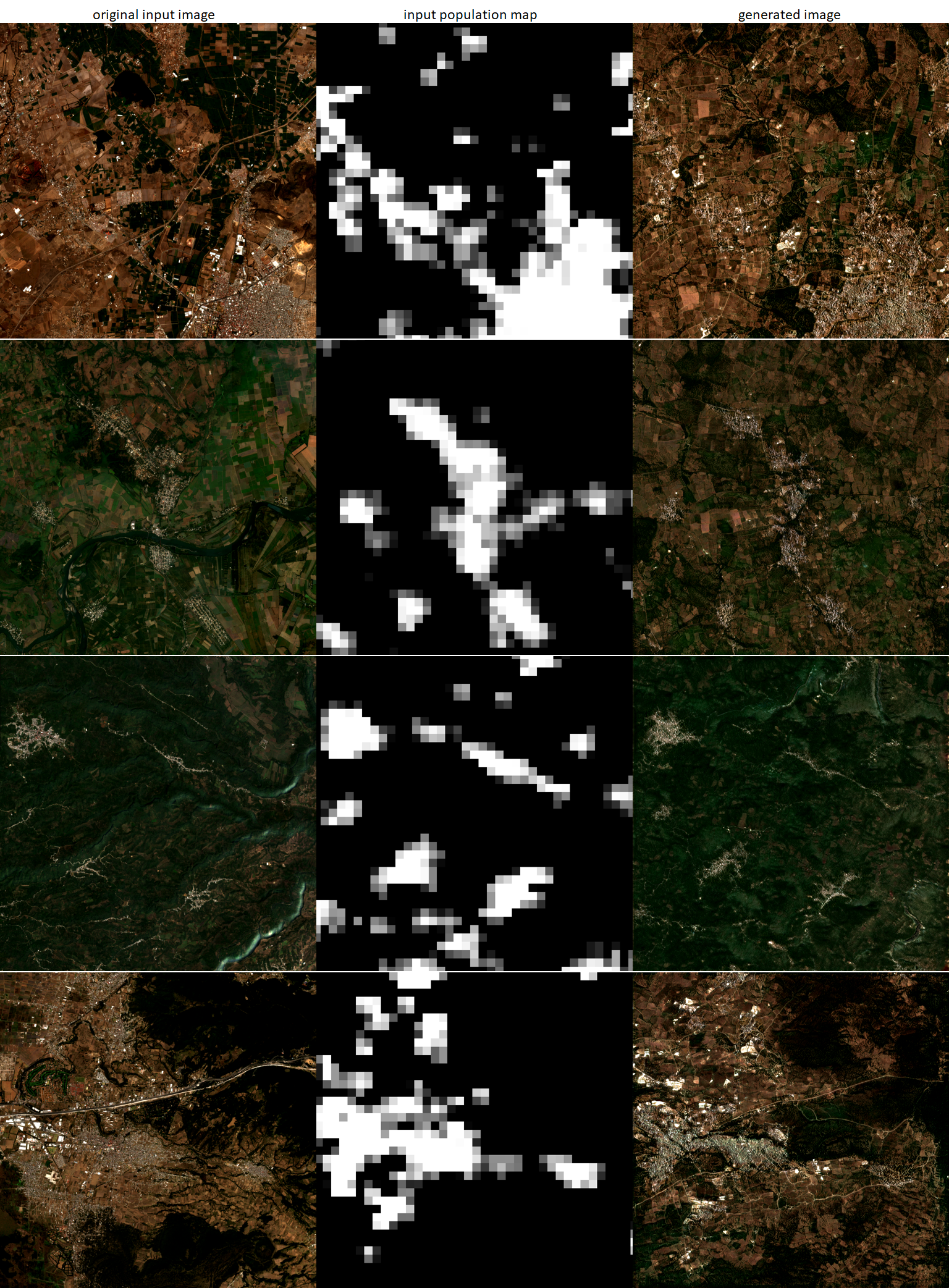}
  \caption{\textbf{Targeted reconstructions} Additional examples}
  \label{reconstructions_apendix}
\end{figure}

\subsubsection{Population pixel difference} \label{pop_pix_dif_appendix}
\begin{figure}[H]
  \centering
  \includegraphics[height = 13cm, width = 13cm]{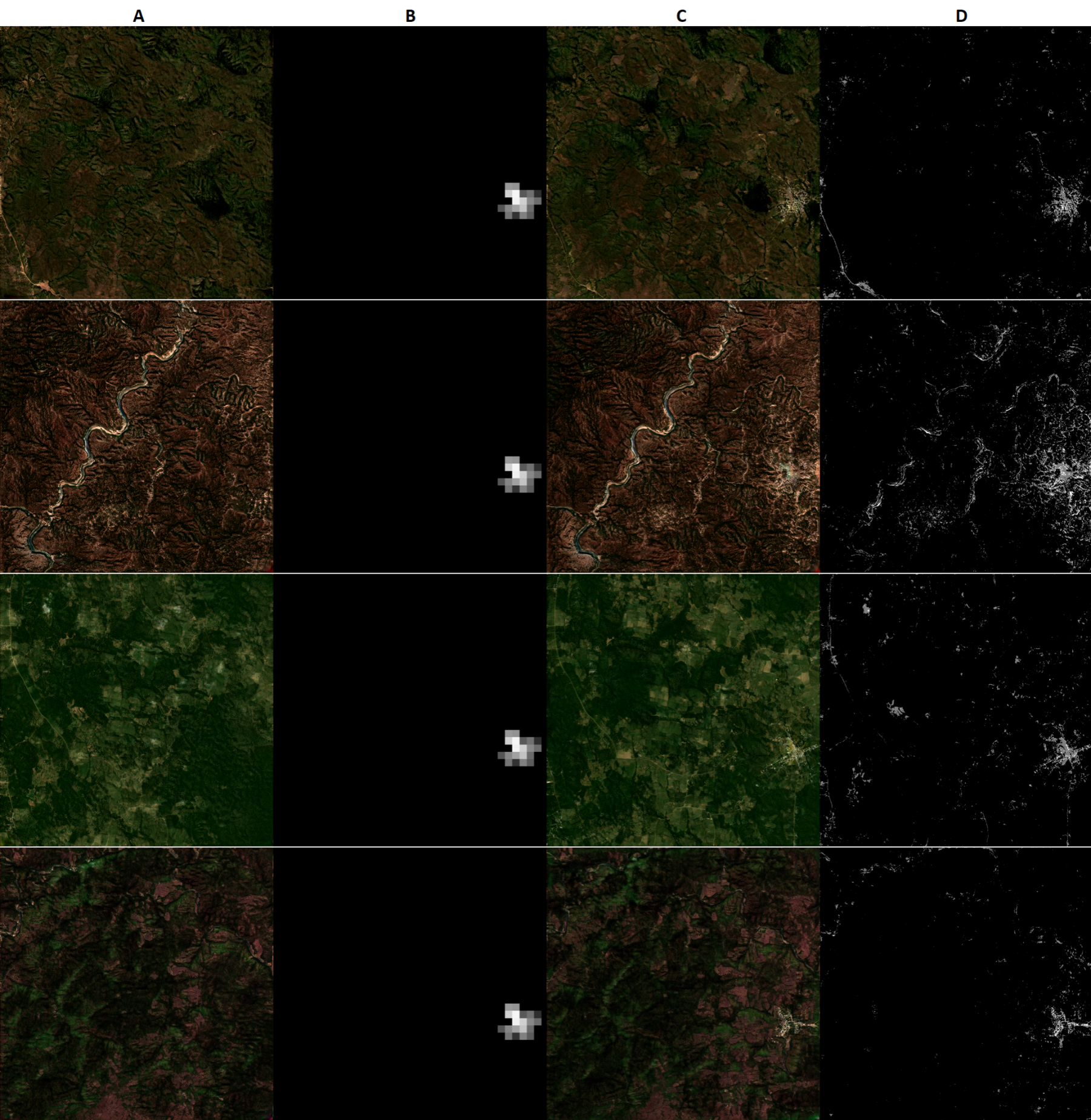}
  \caption{\textbf{Population pixel difference} Panel D contains the pixel difference between the generated images (panels A and C) and clearly reproduces the input population (panel B). Still, speckles of small pixel differences appear throughout the image. This shows that changing the population has some impact globally, suggesting some level of entanglement between the population map and the latent style \emph{w}.}
  \label{pop_diff_ex}
\end{figure}

\subsubsection{Population manipulation}\label{pop_manipulation_appendix}
\begin{figure}[H]
  \centering
  \includegraphics[height = 18cm, width = 7.2cm]{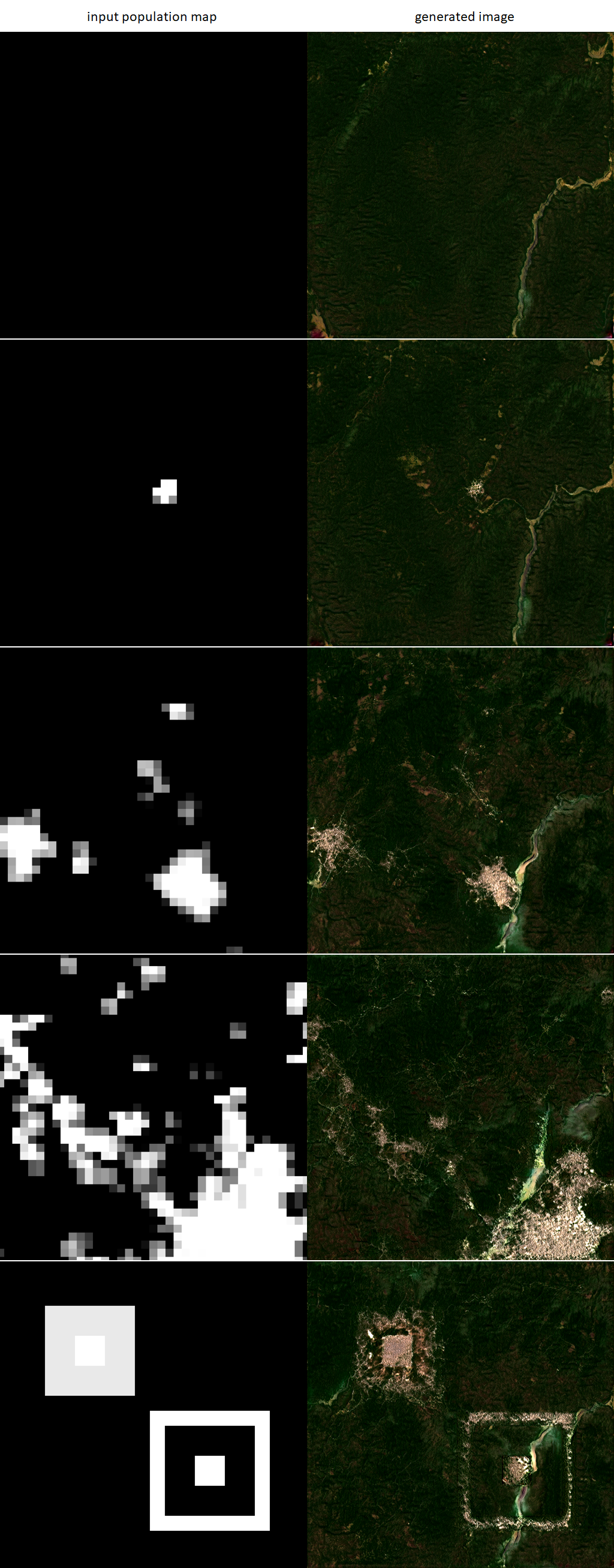}
  \caption{\textbf{Custom population input} We fix the latent style input \emph{w} and change population input. The population is accurately generated, even with unrealistic population distribution in the last row.}
  \label{pop_manipulation}
\end{figure}

\subsubsection{Best and worst reconstructions}\label{best_worst_appendix}
\begin{figure}[H]
  \centering
  \includegraphics[height = 16cm, width = 12cm]{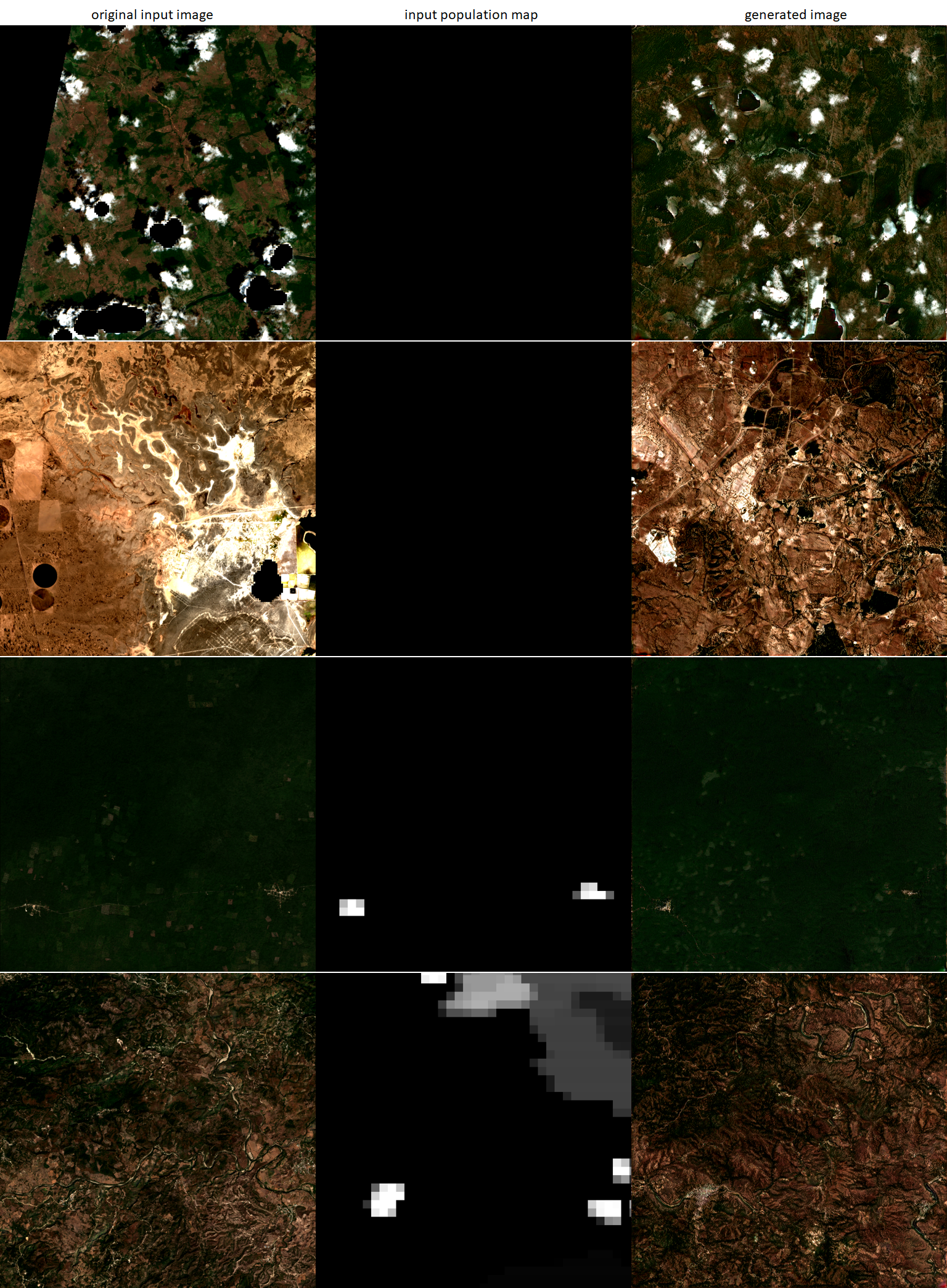}
  \caption{\textbf{Best and worst generations} From top to bottom: semantic worst, pixel worst, pixel best, semantic best}
  \label{best_worst}
\end{figure}

\end{document}